\documentclass[10pt]{article}
\usepackage{multicol}
\usepackage[utf8]{inputenc}
\usepackage[T1]{fontenc}
\usepackage{amsmath,amsfonts,amssymb}
\usepackage{graphicx}
\usepackage[numbers]{natbib}
\usepackage{url}
\usepackage[margin=1.1in]{geometry}
\usepackage{booktabs}
\usepackage{hyperref}

\hypersetup{
    pdftitle={SemanticCite: Citation Verification with AI-Powered Full-Text Analysis and Evidence-Based Reasoning},
    pdfauthor={Sebastian Haan},
    pdfsubject={AI-powered system for verifying citation accuracy through full-text source analysis with transparent, evidence-based reasoning and multi-class classification},
    pdfkeywords={citation verification, reference verification, semantic citation errors, full-text analysis, evidence-based reasoning, explainable AI, AI-assisted research, academic integrity, citation accuracy, multi-class classification, AI hallucination detection, transparent reasoning, source analysis},
    pdfcreator={Sebastian Haan},
    pdfproducer={pdfTeX},
    colorlinks=true,
    linkcolor=blue,
    citecolor=blue,
    urlcolor=blue,
    bookmarksnumbered=true,
    bookmarksopen=true,
    pdfstartview=FitH
}

\title{\raisebox{-0.34\height}{\includegraphics[height=2.1em]
{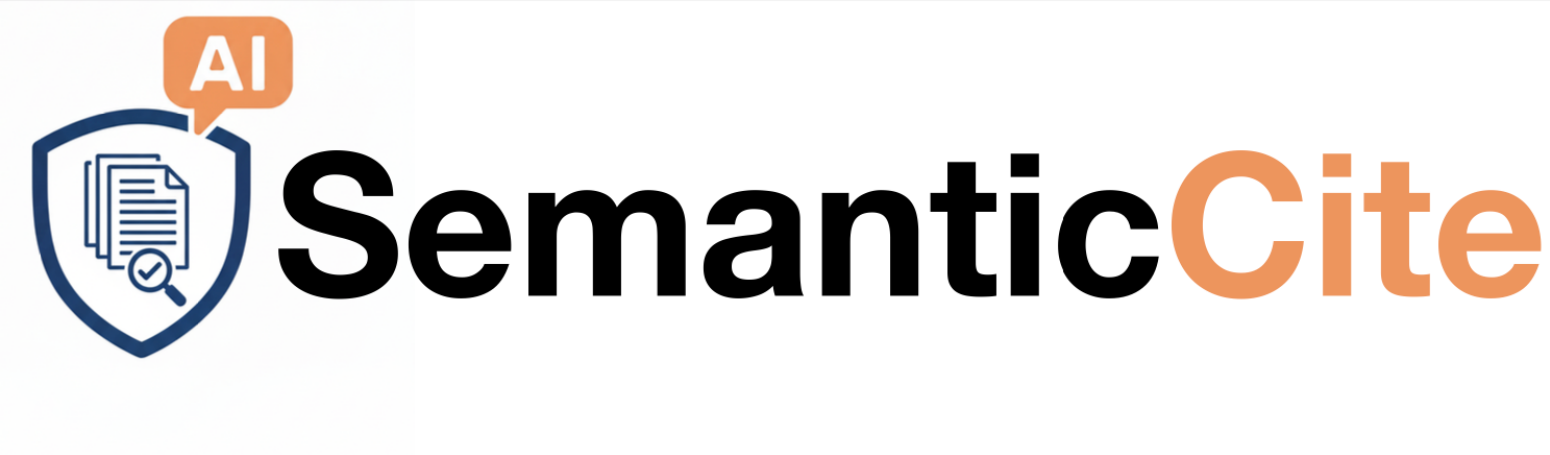}}: \Large\textbf{Citation Verification with AI-Powered Full-Text Analysis and Evidence-Based Reasoning}}

\author{
    Sebastian Haan \\
    The University of Sydney, Australia \\
}

\date{}

\begin{document}

\maketitle
\vspace{-10mm}
\begingroup
\renewcommand\thefootnote{}\footnotetext{Email: sebastian.haan@sydney.edu.au}%
\addtocounter{footnote}{-1}%
\endgroup

\begingroup
\renewcommand\thefootnote{}\footnotetext{Software Repository: https://github.com/sebhaan/semanticcite}%
\addtocounter{footnote}{-1}%
\endgroup

\begin{center}
\href{https://github.com/sebhaan/semanticcite}{%
  \raisebox{-0.1\height}{\includegraphics[height=1em]{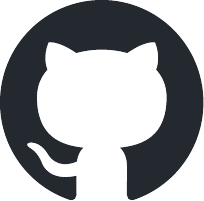}}%
  \hspace{0.3em}{Repository}
}%
  \quad \quad \quad
\href{https://huggingface.co/spaces/sebsigma/semanticcite}{
  \raisebox{-0.25\height}
  {\includegraphics[height=1.5em]{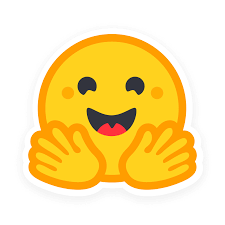}}
  \hspace{0.1em}{Model \& Data}%
}
\end{center}

\begin{abstract}

Effective scientific communication depends on accurate citations that validate sources and guide readers to supporting evidence. Yet academic literature faces mounting challenges: semantic citation errors that misrepresent sources, AI-generated hallucinated references, and traditional citation formats that point to entire papers without indicating which sections substantiate specific claims.
We introduce SemanticCite, an AI-powered system that verifies citation accuracy through full-text source analysis while providing rich contextual information via detailed reasoning and relevant text snippets. Our approach combines multiple retrieval methods with a four-class classification system (Supported, Partially Supported, Unsupported, Uncertain) that captures nuanced claim-source relationships and enables appropriate remedial actions for different error types.
Our experiments show that fine-tuned lightweight language models achieve performance comparable to large commercial systems with significantly lower computational requirements, making large-scale citation verification practically feasible. The system provides transparent, evidence-based explanations that support user understanding and trust.
We contribute a comprehensive dataset of over 1,000 citations with detailed alignments, functional classifications, semantic annotations, and bibliometric metadata across eight disciplines, alongside fine-tuned models and the complete verification framework as open-source software.
SemanticCite addresses critical challenges in research integrity through scalable citation verification, streamlined peer review, and quality control for AI-generated content, providing an open-source foundation for maintaining citation accuracy at scale.

\end{abstract}


\section{Introduction}

\subsection{Problem Statement}

Citations serve as fundamental channels through which scientific knowledge is shared and developed within the academic ecosystem. Despite their critical importance, scholarly literature suffers from a notable prevalence of citation errors that compromise research integrity and quality. 

These errors manifest in several forms: citations supporting claims not made in referenced works, citations omitting critical qualifications present in sources, citations attributing findings to incorrect studies, and selective citation of results whilst ignoring contradictory evidence. Beyond these semantic errors, typographical mistakes in bibliographic information also occur, though digital citation management tools have reduced their frequency. Analysis of prestigious science journals reveals that semantic citation errors remain pervasive, with error rates of 25\% on average \cite{smith2020quotation}.

The implications for research quality are profound. Incorrect citations inadvertently propagate erroneous ideas, contribute to the acceptance of scientific myths, and ultimately erode the credibility of academic scholarship \cite{gotzsche2022citation}. Such inaccuracies can lead to unintended plagiarism, diminished academic standing, and misallocation of valuable research resources.

Manual citation verification during peer review faces substantial limitations. Reviewers are increasingly burdened by overwhelming workloads due to rapid expansion in academic publishing, making it difficult to dedicate sufficient time to meticulous citation evaluation. The lack of standardised review criteria and training contributes to inconsistent evaluations, whilst personal biases can subtly influence outcomes \cite{smith2006peer,tennant2020limitations,mishra2025challenges,sizo2025defining}. 

A pressing new challenge has emerged with AI-generated content, where advanced language models produce convincing but non-existent ``fake citations'' that are extraordinarily difficult to detect manually at scale. Recent studies demonstrate that even advanced models like GPT-5 fabricate citations 39\% of the time when operating without internet access, highlighting the fundamental nature of this problem \cite{gibney2025hallucination}. These hallucinations arise from the statistical nature of large language models, which make predictions by generalising from learned associations, leading to plausible but sometimes incorrect outputs \cite{gibney2025hallucination}. As AI-assisted writing becomes prevalent in academic contexts, traditional verification approaches designed for human authors prove insufficient for detecting AI-generated content that can produce highly plausible but entirely fabricated citations at scale.

Traditional citation presentation provides no mechanism for readers to quickly understand cited work context without manually retrieving and reading entire source documents. Citations merely point to entire papers without indicating specific supporting sections, forcing readers to interrupt their flow, locate sources, search through lengthy documents for relevant passages, and return to the original text. This lack of precise contextual guidance impedes efficient knowledge discovery and creates barriers to thorough fact-checking during both research and peer review processes.

Current automated citation tools remain severely limited across multiple dimensions. Reference management software (EndNote, Zotero, Mendeley) excel at bibliographic formatting but provide no semantic verification capabilities. Existing automated verification systems validate citations against authority databases (Web of Science, PubMed, CrossRef), detecting incorrect metadata or non-existent references, but cannot verify whether citation claims accurately reflect source document content. A reference may exist with correct metadata whilst the citing author mischaracterises the source's actual findings. Similarly, format-checking systems verify style compliance but do not analyse semantic alignment between claims and sources.

Even systems attempting content-level verification typically rely on abstract-only analysis rather than examining complete source documents. This approach misses critical supporting evidence, methodological caveats, limitations discussions, and contradictory findings present in full papers. Important contextual information such as experimental conditions, population constraints, statistical significance details, and author-acknowledged limitations rarely appears in abstracts yet proves essential for accurate citation verification. 

Furthermore, traditional verification approaches employ simple binary schemes (supported\slash unsupported) that fail to capture partial support, where core assertions align with sources but contextual elements are missing. These situations require different remedial actions than completely unsupported claims, yet binary systems cannot distinguish between them. Additionally, automated systems provide classifications without explaining reasoning or presenting supporting evidence, preventing independent validation.

These converging challenges create an urgent need for comprehensive citation verification systems. Pervasive semantic errors, limited verification tools, opaque automated systems, AI-generated hallucinations, and manual verification friction demand solutions that detect errors, explain reasoning, provide actionable guidance, and scale efficiently whilst maintaining the transparency essential for scholarly work.

\subsection{Research Objectives}

This work addresses these fundamental limitations through four primary objectives. First, we develop an automated full-text citation verification system that transcends surface-level bibliographic checks to perform deep semantic analysis of citation content against complete source documents. Second, we create a nuanced 4-class classification scheme (Supported, Partially Supported, Unsupported, Uncertain) that captures the complexity of citation-reference relationships beyond traditional binary approaches, enabling appropriate remedial actions for different types of citation errors. Third, we demonstrate that fine-tuned lightweight models can achieve competitive performance with large commercial language models whilst requiring significantly reduced computational resources. Fourth, we implement transparent evidence-based reasoning that enables users to understand and independently validate system decisions through relevant text snippets extracted from source documents and detailed explanations of classification rationale. This transparency is essential for maintaining the verifiability and trust required in academic contexts, addressing the opacity that has plagued previous automated verification systems.

\subsection{Contributions}

This paper introduces SemanticCite, the first comprehensive AI-powered framework for automated full-text citation verification. Unlike existing approaches that rely on abstract-level analysis or binary classification schemes, our system performs deep semantic analysis of complete source documents using a 4-class taxonomy that captures the nuanced relationships between citations and their references. SemanticCite bridges the gap between automated detection and actionable remediation by providing detailed evidence-based reasoning that enables authors and reviewers to understand and address citation inaccuracies systematically. This represents a fundamental advance from simple error detection to comprehensive citation quality assessment with practical guidance for manuscript improvement. By combining hybrid retrieval and reranking techniques with fine-tuned language models, SemanticCite enables thorough citation analysis while maintaining computational efficiency, making sophisticated verification accessible to individual researchers and institutions regardless of computational resources.

We contribute three key resources spanning data, models, and implementation. Our dataset comprises 1,111 analysed citations across 8 academic disciplines, including extracted claims, 4-class semantic classifications, reasoning explanations, confidence scores, and ranked evidence snippets. We release fine-tuned Qwen3 models (1.7B and 4B parameters) trained for both citation preprocessing and semantic classification tasks that achieve competitive performance with commercial systems  while requiring significantly fewer computational resources. We open-source the SemanticCite framework code, including claim extraction pipelines, hybrid retrieval systems, and classification method.

\section{Related Work}

\subsection{Citation Analysis and Verification}

Traditional citation analysis has primarily focused on bibliometric approaches using citation counts as indicators of research impact, failing to capture semantic relationships between citing statements and referenced sources. The field of automated fact-checking has provided important foundations for citation verification. Wadden et al.~(2020) \cite{wadden2020fact} introduced scientific claim verification with their SciFact dataset comprising 1.4K expert-written claims paired with abstracts annotated for veracity, demonstrating that domain adaptation significantly improves performance versus general-purpose fact-checking models.

However, existing systems face significant limitations. Most rely on abstract-level analysis rather than full-text examination, potentially missing crucial evidence within complete documents. Binary classification schemes (SUPPORTS/REFUTES) fail to capture nuanced citation-reference relationships \cite{wadden2020fact}. The emergence of AI-generated content has introduced new challenges, with advanced language models producing convincing but non-existent citations. Whilst tools like CiteSure \cite{citesure2024} address this specific problem and systems like Morressier's Integrity Manager detect various misconduct indicators, these specialised tools focus on specific error types rather than comprehensive citation verification \cite{sae2024research}.

Some existing platforms analyse citation usage within citing documents, extracting surrounding context and classifying whether the citing statement supports or contradicts the reference generally \cite{nicholson2021scite}. However, these systems examine the directionality in reverse, i.e., they assess whether the citation supports the reference rather than whether the source document supports the specific citation claim. They cannot verify if the referenced source actually contains the evidence the citing author asserts, leaving the fundamental question of claim accuracy unaddressed.

\subsection{Multi-class Citation Classification in NLP}

The evolution from binary to multi-class classification reflects growing recognition that real-world relationships exist on spectra rather than simple dichotomies. Multi-class approaches offer significant advantages by capturing degrees of alignment rather than forcing binary decisions. Systems employing classifications such as ``supported,'' ``partially supported,'' ``unsupported,'' and ``uncertain'' provide granular assessments that better reflect scientific discourse complexity \cite{haan2025developing}.

The Jurgens et al. (2018) taxonomy \cite{jurgens2018measuring} represents a notable contribution, defining six categories (Background, Extends, Uses, Motivation, Compare/Contrast, Future work) for citation intent classification using the ACL-ARC dataset of 1,941 expert-annotated instances. 
Subsequent research expanded available data with the SciCite dataset, over five times larger than ACL-ARC, simplifying taxonomy to three categories (Background, Method, Result) whilst achieving superior performance with pre-trained language models like XLNet (88.93\% F1 score) \cite{mercier2020impactcite}.

The diversity of taxonomies across datasets indicates lack of universal agreement on citation function definitions. This fragmentation, coupled with recognition that citations may serve multiple purposes, suggests current single-label classification may oversimplify relationships, necessitating flexible frameworks supporting customisable taxonomies and multi-label classification \cite{abu-jbara2013,lauscher2021multicite}.

\subsection{Retrieval-Augmented Generation}

Retrieval-augmented generation (RAG) has emerged as a powerful paradigm for fact-checking applications, combining information retrieval with language model generation capabilities \cite{lewis2020retrieval,gao2023retrieval}. In citation verification contexts, RAG enables systems to retrieve relevant evidence from source documents and generate informed assessments of citation accuracy. Traditional RAG approaches face limitations when applied to academic texts due to their complexity and specialised terminology \cite{li2024budget}.

Recent advances in reverse RAG methodologies show promise for citation verification by first extracting or summarising information, then linking each data point back to source documents through systematic fact verification. This approach ensures every claim is traceable to original sources, addressing hallucination problems in standard generative models \cite{rajpurkar2024claimify,metropolitansky2025towards}.

Hybrid retrieval approaches combining dense semantic search with sparse keyword matching have demonstrated superior performance over individual methods \cite{arivazhagan2023,sawarkar2024blended,sultania2024domain}. Dense retrieval using vector embeddings captures semantic relationships, whilst sparse methods like BM25 (Robertson \& Zaragoza, 2009) ensure exact term correspondence. Neural reranking using cross-encoders further optimises candidate ordering based on query-document relevance, providing robust foundations for evidence-based citation verification systems.

\section{Methodology}

\subsection{System Architecture}

\begin{figure*}[!t]
\centering
\includegraphics[width=\textwidth]{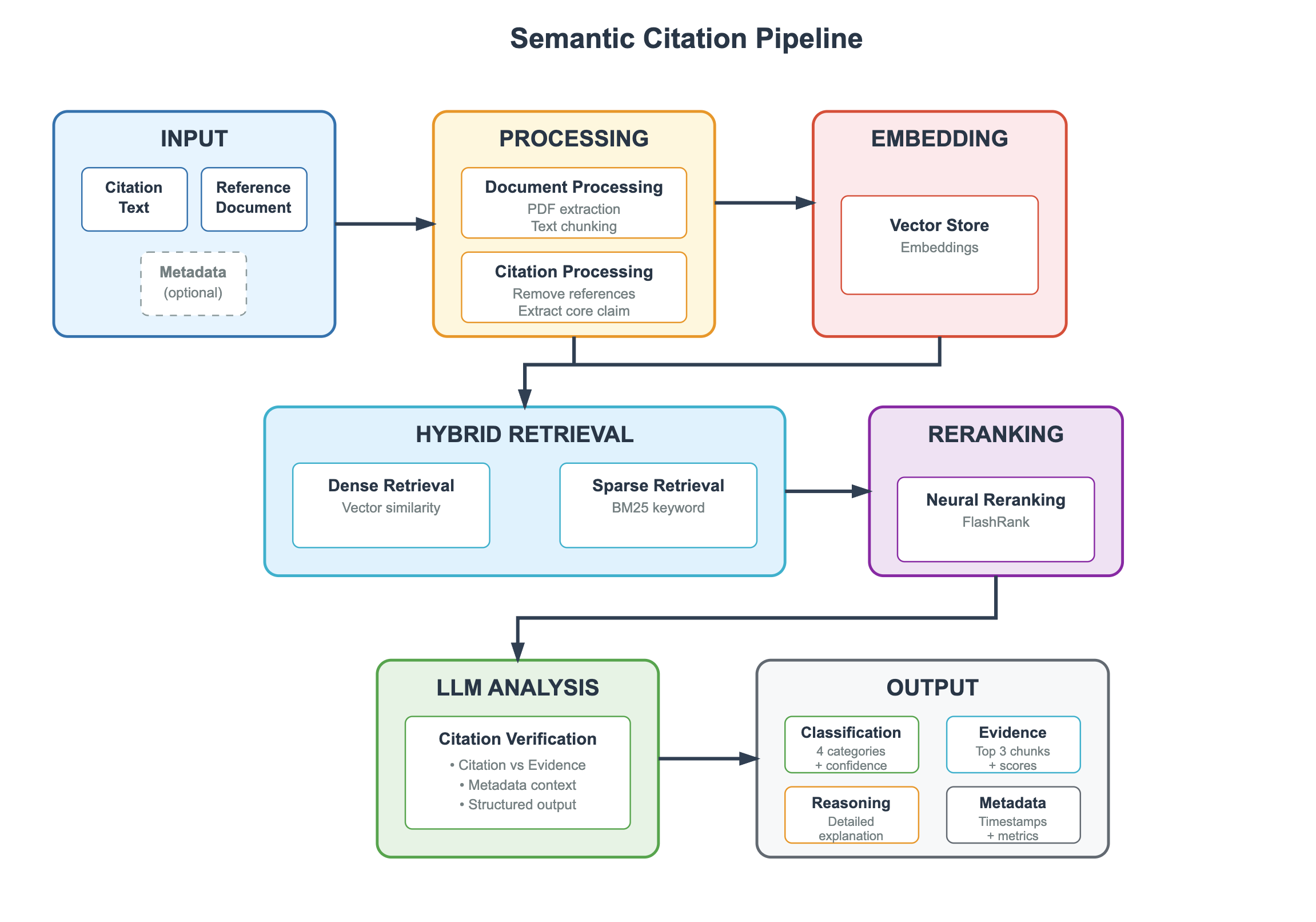}
\caption{Semantic Citation Verification Pipeline: A multi-stage automated system for citation verification combining document processing (PDF extraction and text chunking), vector embedding storage, hybrid retrieval using both dense semantic similarity and sparse BM25 keyword matching, neural reranking with FlashRank, and LLM-based analysis. The pipeline outputs a classification result, supporting evidence, detailed reasoning, and confidence score for each citation verification task.}
\label{fig:pipeline}
\end{figure*}

SemanticCite employs a four-stage pipeline that combines hybrid retrieval-augmented generation with large language models for automated citation verification. The system operates through sequential stages: text preprocessing, hybrid retrieval, neural reranking, and LLM-based analysis, implemented using the LangChain framework with support for multiple providers. Figure~\ref{fig:pipeline} provides an overview of the complete citation verification pipeline, illustrating the flow from input processing through to structured output generation.

The citation verification system takes three inputs: citation text, reference document, and optional metadata. Then it produces a classification label, confidence score, supporting evidence, and detailed reasoning. The system processes reference documents through text extraction and chunking, whilst citations undergo preprocessing to remove attribution markers and extract core factual claims.

Document processing begins with text extraction using PyMuPDF\footnote{PyMuPDF Documentation: https://github.com/pymupdf/PyMuPDF} for PDF files, followed by recursive text splitting into 512-character chunks with 50-character overlap. The chunking strategy employs hierarchical separators including paragraph breaks, sentences, punctuation, and whitespace to maintain semantic coherence whilst ensuring manageable sizes for retrieval operations.

Citation preprocessing transforms raw citations into standardised claims by removing reference markers, author attributions, and publication identifiers whilst preserving numerical values and factual assertions. This transformation converts author-centric statements to fact-centric assertions suitable for independent verification. For example, ``Smith et al. (2020) found that exercise reduces cardiovascular risk by 30\%'' becomes ``Exercise reduces cardiovascular risk by 30\%''.

The system supports flexible configuration of LLM providers including OpenAI, Anthropic Claude, Google Gemini, and local models via LiteLLM\footnote{LiteLLM Documentation: https://docs.litellm.ai/docs/}. Similarly, embedding models can be configured across OpenAI, local SentenceTransformers, or custom endpoints, enabling deployment across diverse computational environments whilst maintaining consistent performance. Temperature settings are controlled for reproducible outputs, with deterministic citation processing using zero temperature.

\subsection{Classification Scheme}

\begin{figure*}[!t]
\centering
\includegraphics[width=\textwidth]{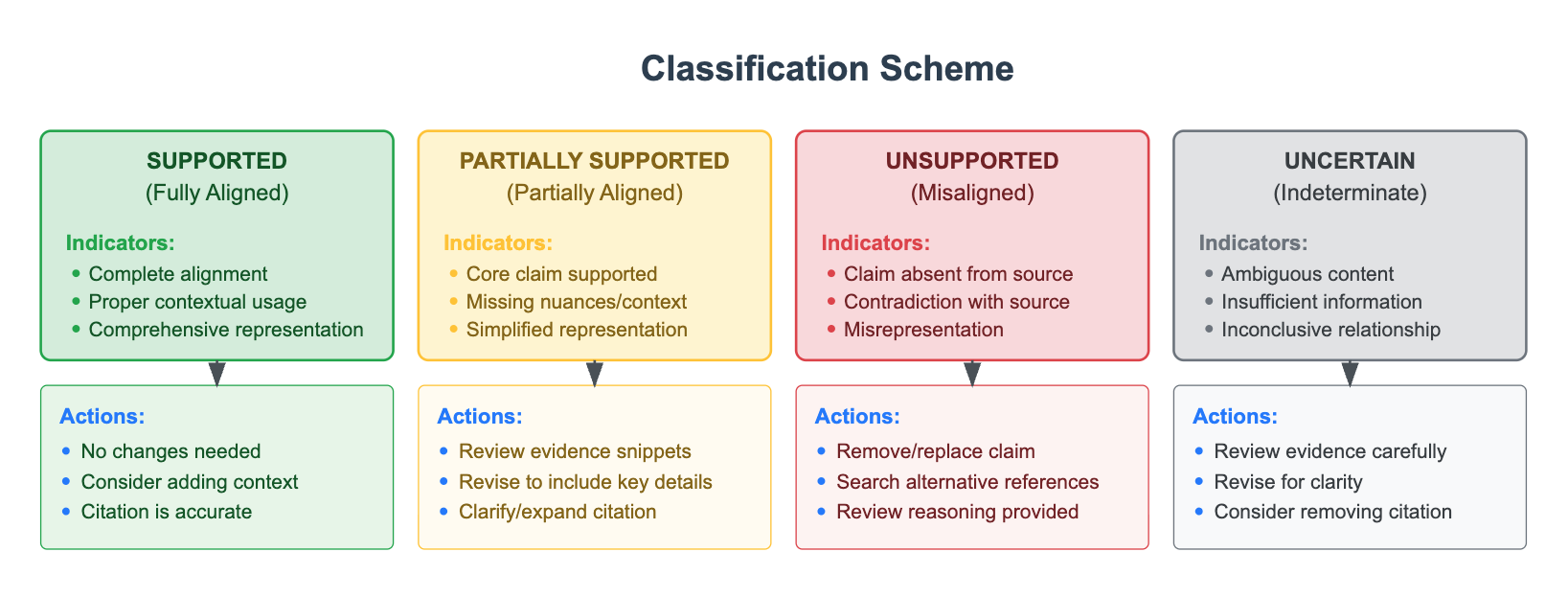}
\caption{Four-Category Classification Scheme for Source-Claim Alignment Assessment}
\label{fig:classification}
\end{figure*}

Our 4-class taxonomy represents a fundamental contribution to citation verification methodology, providing nuanced assessment beyond traditional binary approaches. The classification scheme follows a logical progression from fully supported to unsupported citations, with an uncertainty class capturing ambiguous cases where clear determination cannot be made. Figure~\ref{fig:classification} illustrates the complete classification framework with actionable guidance for each category.

The development of this taxonomy addresses critical limitations in existing binary classification systems. Traditional SUPPORTS/REFUTES schemes, whilst computationally simpler, fail to capture the complexity of citation-reference relationships commonly encountered in academic literature. Many citations exist in intermediate states requiring different remedial actions, making binary classification insufficient for practical citation verification applications.

\subsubsection{Detailed Classification Categories}

\textbf{SUPPORTED (Fully Aligned):} Citations demonstrate complete alignment between claims and source material, maintaining proper context and comprehensive representation. These citations accurately convey the referenced content without oversimplification, missing context, or distortion of meaning. The core indicators include complete factual alignment, appropriate contextual usage, and comprehensive representation of source nuances. For instance, a citation stating ``The human genome contains approximately 20,000 protein-coding genes'' fully aligns with source material stating ``Recent genomic studies have confirmed that the human genome encodes about 20,000 protein-coding genes.'' Such citations receive high confidence scores and require no modifications, though users may optionally add additional context from provided evidence snippets.

\textbf{PARTIALLY SUPPORTED (Partially Aligned):} Core claims are present in source material but lack critical nuances, contextual information, or important qualifications, leading to incomplete representation. These citations contain accurate fundamental assertions but oversimplify complex findings or omit essential caveats present in the source. Key indicators include core claim support with missing minor nuances, simplified representation, or incomplete contextual information. An example would be citing ``Exercise guarantees a 50\% reduction in cardiovascular disease risk'' when the source states ``Regular exercise has been shown to reduce cardiovascular disease risk by up to 50\% in some studies, though results can vary.'' The citation captures the core finding but overstates certainty and omits variability. Such citations receive medium confidence scores and require minor revisions to include qualifying details and expand scope for complete accuracy.

\textbf{UNSUPPORTED (Misaligned):} Claims are absent from, contradicted by, or significantly misrepresent source material. These citations demonstrate fundamental disconnection between stated claims and referenced content, ranging from complete absence of supporting evidence to direct contradiction of source findings. Critical indicators include claims absent from source material, direct contradiction between claims and evidence, or significant misrepresentation of source content. For example, claiming ``Drinking coffee cures cancer'' when the source states ``Some studies suggest compounds in coffee may lower risk of certain cancers, but evidence is not conclusive'' represents serious misrepresentation. Such citations receive low confidence scores and require major revision, removal, or alternative reference identification.

\textbf{UNCERTAIN (Indeterminate):} Ambiguous relationships exist where insufficient evidence, unclear claim-reference relationships, or inadequate verification context prevent definitive classification. These cases often involve vague claims, complex source material with multiple interpretations, or insufficient contextual information for reliable assessment. Key indicators include ambiguous claim or source content, insufficient contextual information, or inconclusive claim-reference relationships. An example might be citing ``Vitamin D supplementation prevents all cases of osteoporosis'' when the source discusses ``varying degrees of effectiveness depending on factors such as age, dosage, and baseline deficiency levels.'' The broad generalisation in the claim combined with nuanced source findings creates ambiguity requiring additional clarification. Such citations warrant careful review of evidence snippets to identify ambiguity sources.

\subsubsection{Actionable User Guidance Framework}

The classification scheme provides specific, actionable guidance aligned with potential user responses:
\begin{itemize}
\item SUPPORTED citations receive confirmation that no changes are needed, with optional suggestions for additional context enhancement.
\item PARTIALLY SUPPORTED citations receive specific guidance on reviewing evidence snippets to identify missing nuances and revising content to include qualifying details.
\item UNSUPPORTED citations prompt users to remove inaccurate claims, replace them with evidence-based statements, or seek alternative supporting references.
\item UNCERTAIN classifications guide users to carefully review evidence snippets, revise citations to include necessary contextual details, or consider citation removal if relationships remain unclear.
\end{itemize}

This structured guidance framework transforms abstract classification labels into concrete editorial actions, bridging the gap between automated analysis and practical manuscript improvement. The approach recognises that different classification outcomes require different remedial strategies, moving beyond simple accuracy scores to provide constructive improvement pathways.

\subsubsection{Advantages Over Binary Classification}

The multi-class approach offers several critical advantages over binary schemes. Binary classification forces complex citation-reference relationships into oversimplified dichotomies, failing to distinguish between minor revisions and major corrections. Our taxonomy captures nuanced relationships requiring different remedial actions, providing more precise feedback for authors and reviewers.

Furthermore, the inclusion of an UNCERTAIN category acknowledges inherent limitations in automated analysis, providing transparent handling of ambiguous cases rather than forcing potentially incorrect binary decisions. This approach enhances system trustworthiness by explicitly flagging cases requiring human judgment whilst maintaining high confidence in clear-cut classifications.

The taxonomy also aligns with real-world editorial workflows where citations often require incremental improvements rather than complete acceptance or rejection. By providing graduated classification levels, the system supports iterative manuscript refinement and targeted quality improvement efforts.

\subsection{Hybrid Retrieval System}

The retrieval architecture combines dense semantic search with sparse keyword matching to maximise evidence recall whilst maintaining precision. This hybrid approach addresses individual method limitations: dense retrieval captures semantic relationships but may miss exact term matches, whilst sparse retrieval ensures keyword correspondence but lacks semantic understanding.

\textbf{Dense Retrieval} employs vector embeddings stored in ChromaDB using cosine similarity matching. Document chunks are mapped into high-dimensional vector spaces where semantically similar content clusters together. When processing citation claims, the system calculates similarity scores between query vectors and stored document vectors, retrieving the 15 most semantically relevant chunks. This approach excels at capturing conceptual relationships and paraphrased content that might use different terminology whilst conveying similar meaning.

\textbf{Sparse Retrieval} utilises BM25 algorithm for exact term matching, ensuring that citations requiring specific keyword correspondence are properly identified. BM25 considers term frequency, document length normalisation, and inverse document frequency to rank passages based on lexical overlap with citation claims. This method retrieves an additional 15 chunks focusing on exact term matches, complementing semantic retrieval by capturing precise terminology and specific numerical values.

\textbf{Neural Reranking} applies FlashRank cross-encoder optimisation to the combined retrieval results. After deduplication, the hybrid results undergo reranking based on query-document relevance scores computed by BERT-based cross-encoders. This process optimises candidate ordering and selects the top 3 most relevant passages for downstream analysis, ensuring focus on the most pertinent evidence whilst maintaining computational efficiency.

\textbf{Document Chunking Strategy} employs recursive character splitting with carefully chosen parameters to balance granularity with context preservation. The 512-character chunk size with 50-character overlap ensures sufficient context for semantic understanding whilst maintaining computational tractability. Hierarchical separators prioritise natural text boundaries, starting with paragraph breaks and progressively falling back to sentences, punctuation, and finally individual words when necessary.

The hybrid approach demonstrates superior performance compared to individual methods by leveraging complementary strengths. Dense retrieval captures nuanced semantic relationships crucial for academic content, whilst sparse retrieval ensures critical exact matches are not missed. Neural reranking provides final optimisation, resulting in high-quality evidence selection for citation verification.

\subsection{Citation Processing Pipeline}

The processing pipeline transforms raw academic citations into verifiable claims whilst preserving essential factual content. This preprocessing stage is critical for accurate verification as raw citations often contain attribution elements that obscure core factual assertions.

\textbf{Preprocessing Operations} systematically transform citations through several key steps. Reference markers such as brackets, parentheses, and citation numbers are removed to isolate factual content from bibliographic formatting. Author attributions are eliminated whilst preserving substantive claims, converting statements like ``Smith et al. found that X'' to ``X was found'' using passive voice constructions. Numerical values, statistical measures, and specific metrics are carefully preserved as these often represent the most verifiable aspects of citations. The process ensures all claims become independently verifiable statements that can be assessed against source material without requiring external attribution context.

\textbf{Structured Output Parsing} employs LLM-based analysis with explicit schema requirements to ensure consistent and comprehensive results. The system generates structured JSON outputs containing classification labels, detailed reasoning with both summary and point-by-point analysis, confidence scores, and ranked evidence snippets with relevance scores and source locations. This structured approach enables programmatic processing whilst maintaining human interpretability.

The verification prompt carefully incorporates processed claims, top-ranked evidence snippets with relevance scores, and optional metadata for enhanced contextual understanding. Prompts explicitly instruct models to assess factual alignment between citations and source material whilst considering broader study context when metadata is available. Instructions emphasise objective evaluation based on textual evidence rather than general knowledge or assumptions.

\textbf{Evidence Snippet Ranking} applies systematic relevance scoring to prioritise the most pertinent supporting material. Each evidence snippet receives numerical relevance scores indicating how well it addresses the citation claim, enabling users to focus on the most important supporting or contradicting evidence. The system tracks comprehensive metadata including processing timestamps, reference file identifiers, chunk locations, and system version information for complete result provenance and reproducibility analysis.

This systematic approach enables transparent verification where users understand classification rationale through extracted evidence and detailed reasoning, supporting both automated analysis and human review processes. The pipeline's modular design allows for component-level optimisation and adaptation to different document types and verification requirements whilst maintaining consistent output quality.

\subsection{Web Interface and Deployment}

\begin{figure}[!t]
\centering
\includegraphics[width=1.0\textwidth]{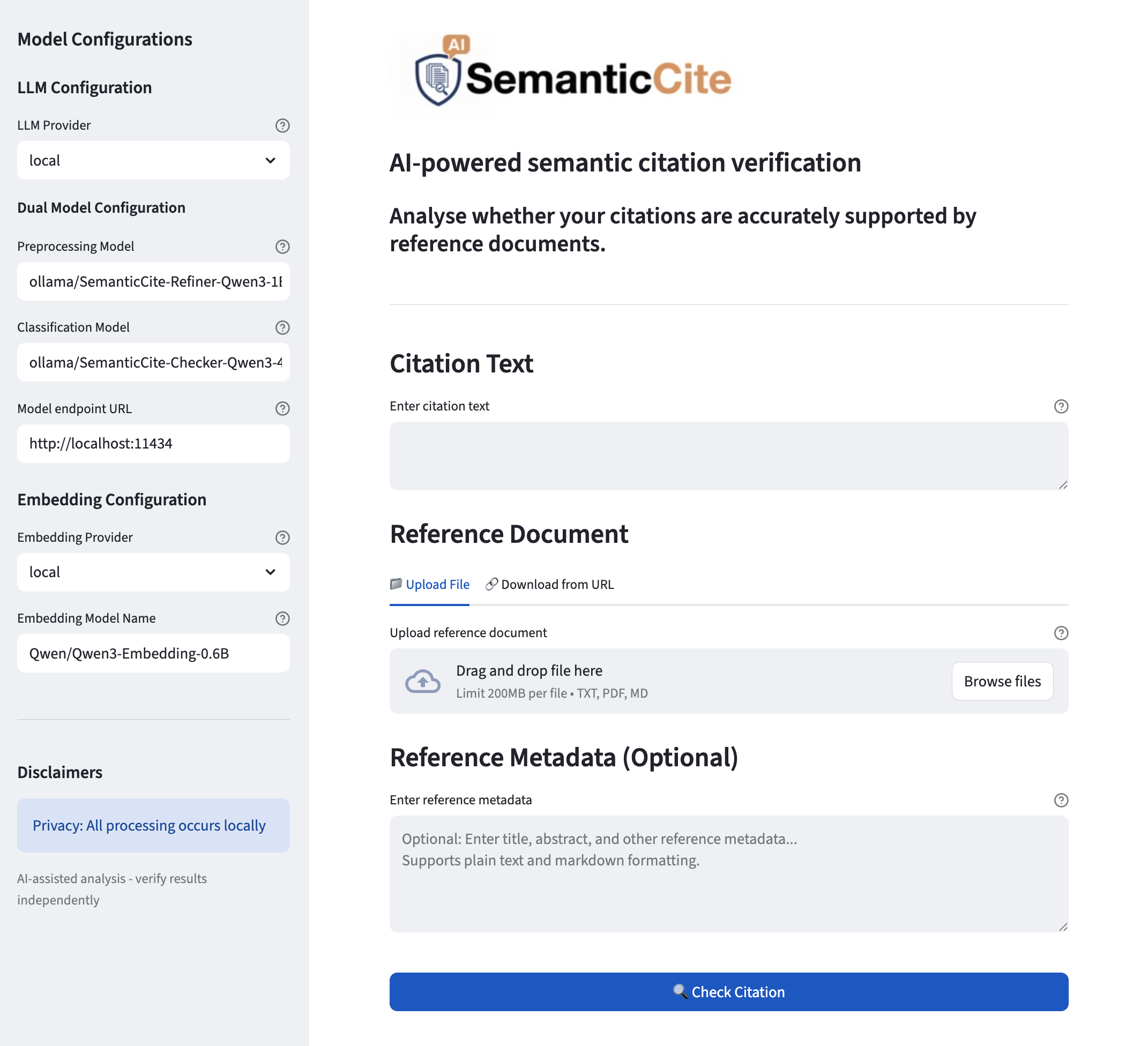}
\caption{SemanticCite web interface showing citation input, reference document upload options (file upload or URL download), optional metadata entry, and configurable model parameters. The interface supports multiple LLM providers and embedding models, enabling flexible deployment across different resource constraints and institutional requirements.}
\label{fig:interface}
\end{figure}

To facilitate widespread adoption and ensure accessibility for researchers without programming expertise, we developed a web-based interface using Streamlit (Figure~\ref{fig:interface}). The interface provides intuitive controls for citation verification, allowing users to input citation text, upload reference documents (PDF, TXT, or Markdown formats), or download references directly from URLs. Users can configure both LLM providers (OpenAI, Anthropic Claude, Google Gemini, or local Ollama models) and embedding models (local SentenceTransformers, OpenAI embeddings, or custom endpoints) through an interactive sidebar, enabling flexible deployment across different institutional resource constraints and privacy requirements. The system displays results including classification labels with color coding, confidence scores, detailed reasoning explanations, and ranked evidence snippets extracted from the reference document. All analysis results can be exported to Markdown format for documentation and integration into research workflows. This implementation demonstrates the practical deployment of the SemanticCite framework whilst maintaining full analytical capabilities and transparency in the verification process.

\section{Model Development and Evaluation}

This section details the development and evaluation of fine-tuned large language models for automated citation verification within the SemanticCite framework. We present our complete experimental pipeline for training specialized citation analysis models, beginning with the systematic construction of a balanced, multi-disciplinary dataset spanning eight academic fields (§4.1). We describe our automated training data generation methodology that leverages LLM-based filtering and annotation to create high-quality datasets for both citation preprocessing and classification tasks (§4.2). The fine-tuning process employs QLoRA techniques with Qwen3 models \cite{qwen2024} of varying scales to develop efficient, specialized systems for citation verification (§4.3). We introduce a weighted evaluation framework specifically designed to assess the ordinal nature of citation support categories (§4.4), and conclude with comprehensive experimental results demonstrating model performance across different scales and computational requirements (§4.5). This experimental framework establishes the effectiveness of fine-tuned language models for citation verification while providing insights into the trade-offs between model complexity, computational efficiency, and task-specific performance in academic text analysis.

\subsection{Data Selection}

\begin{figure*}[!t]
\centering
\includegraphics[width=0.95\textwidth]{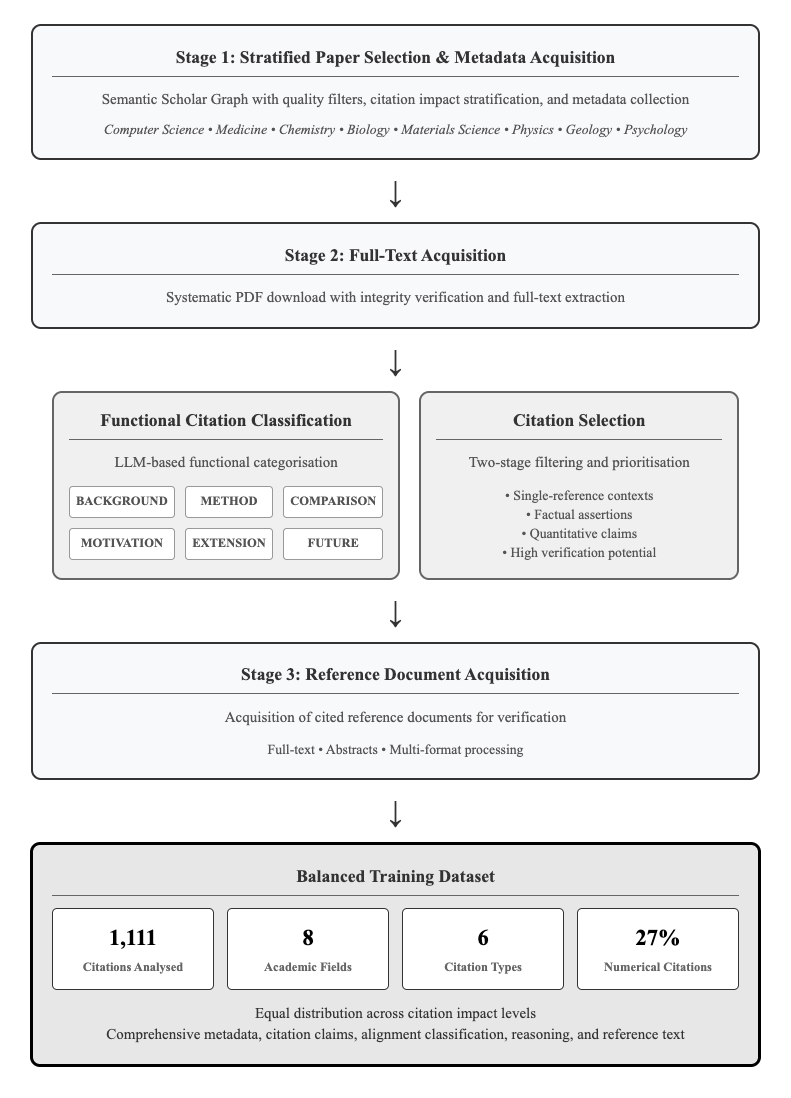}
\caption{Overview of the data selection and processing pipeline for semantic citation verification}
\label{fig:data_selection}
\end{figure*}

We constructed a high-quality training dataset using a multi-stage approach designed to ensure representative coverage across academic disciplines while maintaining rigorous quality standards for bibliometric metadata and full-text content. The resulting corpus targets balanced representation across citation impact levels, disciplinary boundaries, and functional citation categories, providing a robust empirical foundation for training and evaluating citation analysis models. An overview of the data selection process is presented in Figure~\ref{fig:data_selection}.

\subsubsection{Data Collection and Processing Pipeline}

The training dataset was constructed through a systematic three-stage pipeline designed to collect academic papers with diverse citation characteristics across multiple disciplines, targeting a corpus of 4,000 papers with balanced representation across citation impact levels.

\textbf{Stage 1: Stratified Paper Selection}

The collection process employed a stratified sampling approach across eight academic fields (Computer Science, Medicine, Chemistry, Biology, Materials Science, Physics, Geology, and Psychology) using the Semantic Scholar Graph API. This selection ensures comprehensive coverage of distinct disciplinary citation practices and scholarly communication patterns. Papers were filtered according to stringent criteria including publication year (2019-2023), minimum citation count ($\geq 5$), minimum reference count ($\geq 10$), and open access availability. However, open-access designation did not guarantee that all papers were actually available for download. The temporal constraint ensures currency whilst providing sufficient time for citation accumulation, and the citation thresholds guarantee papers with meaningful scholarly impact and comprehensive reference networks.

To ensure balanced representation across citation impact levels, papers were categorised into three citation rate bins (Low, Mid, High) based on field-specific empirically-derived thresholds that account for citations-per-year rates rather than absolute counts. This normalisation approach addresses the substantial variation in citation velocity across academic disciplines. These thresholds varied considerably across disciplines, reflecting domain-specific citation practices—for instance, Medicine exhibited substantially higher thresholds (Mid: 14, High: 31 citations/year) compared to Geology (Mid: 4, High: 7 citations/year). This stratification ensures that the dataset captures papers across the full spectrum of scholarly impact within each discipline, avoiding bias towards highly-cited outliers whilst maintaining representation of influential work.

\textbf{Stage 2: Metadata Acquisition}

For each selected paper, extensive metadata was retrieved including bibliographic information, abstracts, author details, venue information, external identifiers, and complete citation and reference networks. This metadata collection provides essential context for understanding citation patterns and enables sophisticated analysis of citation behaviour across different paper characteristics. Quality assurance measures excluded papers lacking valid open access PDF URLs or failing PDF format validation, ensuring accessibility of full-text content essential for detailed citation analysis.

\textbf{Stage 3: Full-Text Document Acquisition}

The final stage involved systematic download of PDF documents using validated open access URLs. Each download was verified for successful completion and proper PDF format before inclusion in the dataset, with verification including file size validation, PDF structure integrity checks, and preliminary text extraction testing. This methodology yielded a balanced, high-quality corpus of approximately 500 papers per academic field, with equal distribution across citation impact categories. The approach ensures representative coverage across academic disciplines and citation patterns whilst maintaining access to both structured bibliometric data and full-text content, providing a robust foundation for training citation analysis models.

\subsubsection{Citation Function Classification Methodology}

\textbf{Classification Framework}

The citation classification system employs a systematic approach to categorise citation functions within academic texts, building upon the established taxonomy of Jurgens et al. (2018) \cite{jurgens2018measuring}. The methodology implements a six-category schema designed to capture the primary functional roles of citations in scholarly discourse. This approach balances thorough coverage of citation purposes with tractable computational requirements for automated classification.

\begin{itemize}
\item BACKGROUND citations establish the general field, provide contextual information, or reference foundational knowledge essential for understanding the current work.
\item METHOD citations reference specific algorithms, techniques, datasets, software, or experimental procedures employed or adapted in the current work, representing instrumental citations that directly inform methodological choices.
\item RESULT\_COMPARISON citations directly compare or contrast findings, results, or conclusions with those of the cited work, often appearing in results and discussion sections where authors evaluate their findings against existing evidence.
\item MOTIVATION citations indicate that the cited work directly inspired the research problem or provided primary motivation for the current study, representing foundational influences on research direction.
\item EXTENSION citations signal that the current work builds upon, extends, improves, or adapts methods or ideas from the cited work, indicating direct intellectual lineage.
\item FUTURE\_WORK citations mention prior work specifically in the context of suggested research directions or open problems, typically appearing in concluding sections where authors outline potential developments.
\end{itemize}

\textbf{Dual-Dimensional Classification Approach}

The classification system operates along two complementary dimensions to capture both functional and quantitative aspects of citation usage. The primary dimension assigns each citation context to one of the six functional categories through systematic analysis of the surrounding textual context, considering both immediate sentence-level context and broader paragraph-level discourse patterns. The secondary dimension employs binary classification to identify whether the citation context explicitly involves numerical or quantitative information—including percentages, specific values, metrics, performance scores, or statistical data—directly related to the claim or comparison involving the cited work. This dual-dimensional approach recognises that quantitative citations often require different verification strategies compared to qualitative references, as numerical claims demand precise fact-checking against source documents whilst qualitative citations may require more nuanced semantic analysis. The quantitative dimension proves particularly valuable for identifying citations that make specific empirical claims requiring direct verification against source data.

\textbf{Computational Implementation}

The classification process uses large language models (LLMs) with structured prompting to ensure consistency and reproducibility across the entire dataset. Citation contexts are processed in configurable batches to optimise computational efficiency whilst maintaining classification accuracy. Each classification decision is accompanied by a concise rationale explaining the categorical assignment, providing transparency and enabling quality assessment of the automated classification process. The methodology incorporates error handling and validation mechanisms, including verification of output format compliance, categorical validity checks, and fallback procedures for problematic cases.

\subsubsection{Citation Selection Methodology}

To generate high-quality training data for fine-tuning large language models on citation analysis tasks, we implemented a two-stage automated selection pipeline using LLM-based filtering and ranking designed to identify citation instances most suitable for verification analysis.

The first stage applied strict filtering criteria to identify valid citation instances, requiring that: (1) the context contained only a single reference rather than multiple citations, ensuring unambiguous attribution and avoiding complications from citation clustering; (2) the context represented an actual citation sentence rather than a bibliography entry, maintaining focus on in-text scholarly discourse; and (3) the citation text was properly embedded within the contextual passage, ensuring sufficient surrounding context for meaningful analysis.

From the filtered candidates, a second LLM stage selected the top 5 citations per paper that warranted verification analysis. The selection prioritised citations containing specific claims, numerical statements, factual assertions, and substantial content that could be meaningfully verified against their source references, rather than brief mentions or general acknowledgments. This prioritisation ensures that the resulting training data focuses on citations with sufficient semantic content to support meaningful verification tasks. Files with fewer than 3 citations were excluded to maintain dataset quality and ensure sufficient citation diversity within each source document. This systematic approach ensured the selection of citation instances most suitable for training models to assess citation-reference alignment and factual accuracy.

\subsection{Training Data Generation}

Training datasets are generated for two distinct model types addressing different aspects of citation analysis. The Citation Preprocessing Model transforms raw citation statements into cleaned claims through systematic processing, creating input-output pairs mapping original citations to their standardised, processed forms. This preprocessing removes reference markers, author attributions, and publication identifiers whilst preserving numerical values and factual assertions, converting author-centric statements to fact-centric representations suitable for verification.

The Classification Model generates training examples for citation support classification through a more complex pipeline involving: (1) preparing vector stores from reference documents using hybrid retrieval methods, (2) extracting semantically relevant text chunks through embedding-based similarity matching using Qwen3-Embedding-0.6B, and (3) generating structured outputs containing the four classification labels (SUPPORTED, PARTIALLY SUPPORTED, UNSUPPORTED, UNCERTAIN), detailed reasoning, and confidence scores.

Ground truth annotations were generated using GPT-4.1 (gpt-4.1-2025-04-14) with carefully designed prompts ensuring consistent and reliable classifications. The citation preprocessing employed prompts instructing the model to remove reference markers, convert author-centred statements to fact-centred assertions, maintain numerical values, and ensure claims stand alone as verifiable statements.

For classification analysis, GPT-4 was provided with processed citations, reference document metadata, and relevant text snippets, then instructed to classify citations using our 4-class taxonomy. The structured output schema required classification labels, reasoning dictionaries containing summary and detailed analysis, and numerical confidence ratings (0.0-1.0).

\subsection{Model Fine-tuning}

The fine-tuning methodology employed QLoRA (Quantised Low-Rank Adaptation) techniques with Qwen3 language models to develop specialised systems for citation verification.

\textbf{Model Architecture and Quantisation}

The system used Qwen3 models in three configurations (1.7B, 4B, and 8B parameters) with 4-bit quantisation via BitsAndBytes to enable efficient training on consumer hardware. The QLoRA configuration employed a rank (r) of 16 with an alpha value of 32, targeting attention projection layers (q\_proj, k\_proj, v\_proj, o\_proj) and feed-forward network components (gate\_proj, up\_proj, down\_proj). Gradient checkpointing through Unsloth optimisation reduced memory consumption whilst maintaining training stability, enabling deployment across diverse computational environments.

\textbf{Training Data Preparation and Configuration}

Three distinct training datasets were prepared for different citation analysis tasks. The Preprocessing Dataset contained 1,111 examples for text simplification and paraphrasing tasks, with input-output pairs averaging 49.2 and 38.8 tokens respectively. The Classification Dataset comprised 1,111 examples for citation verification tasks, featuring significantly longer sequences (average 810.9 input tokens, 178.5 output tokens) due to inclusion of reference snippets and metadata.

The training protocol implemented a conservative approach with 2 epochs, a learning rate of $2 \times 10^{-4}$, and AdamW optimisation with 8-bit precision. Batch sizes were dynamically adjusted based on GPU memory (1-4 per device) with gradient accumulation steps (2-8) to maintain effective batch sizes. The system employed linear learning rate scheduling with 5\% warmup ratio and 1\% weight decay for regularisation.

\textbf{Task-Specific Training Approaches}

Text Preprocessing Models were trained with sequence lengths of 1024 tokens while the Citation Classification Models used an extended sequence lengths (4608 tokens). Structured Output Models generated JSON-formatted responses containing classification labels, detailed reasoning, and confidence scores (0.0-1.0), enabling programmatic integration and uncertainty quantification. Training data underwent systematic preprocessing whereby complex nested JSON structures were flattened into coherent text sequences, with task-specific instructions integrated using Alpaca formatting templates to create standardised prompt structures.

Models were exported in multiple formats including LoRA adapters, merged weights, and GGUF quantised versions for deployment flexibility. The system maintained reproducibility through fixed random seeds and deterministic training procedures, with data splitting following a 90:10 train-test partition using stratified sampling.

\subsection{Evaluation Methodology}

Our classification task involves predicting four categories with natural ordinal structure: SUPPORTED (3), PARTIALLY SUPPORTED (2), UNSUPPORTED (1), and UNCERTAIN (0). Unlike traditional multi-class evaluation that treats all misclassifications equally, we implement a weighted evaluation framework that penalizes errors proportionally to their semantic distance.

\subsubsection{Weighted Evaluation Framework}

We define a weight matrix where misclassification penalty equals the absolute difference between ordinal values: $W(i,j) = |v(i) - v(j)|$. This ensures that extreme errors (e.g., SUPPORTED $\to$ UNCERTAIN, weight = 3) receive higher penalties than adjacent-level misclassifications (e.g., SUPPORTED $\to$ PARTIALLY SUPPORTED, weight = 1).

Weighted accuracy is calculated as:
\begin{equation}
\text{ACC}_{\text{weighted}} = 1 - \frac{\sum_{i,j} W(i,j) \times \text{CM}(i,j)}{N \times \max(W)}
\end{equation}

where CM$(i,j)$ represents confusion matrix entries, $N$ is total predictions, and $\max(W) = 3$. We also calculate ordinal Mean Absolute Error (MAE) treating classes as ordered variables, providing intuitive 0-1 scale performance assessment.

Our evaluation combines standard metrics (accuracy, F1-scores, Cohen's $\kappa$) with ordinal-aware measures (weighted accuracy, ordinal MAE) to capture both traditional classification performance and task-specific semantic alignment.

\subsubsection{Text Generation Quality Metrics}

Beyond classification accuracy, we assess text generation quality using character-level similarity and length calibration metrics. Character similarity is calculated using Jaccard similarity at the character level, measuring the overlap of unique characters between predictions and ground truth texts. Specifically, we convert both texts to lowercase, remove spaces, create sets of unique characters, and calculate Jaccard similarity as the intersection divided by the union of character sets. This metric captures lexical overlap and vocabulary consistency regardless of word order, providing insight into how well models preserve semantic content during text generation. Length difference measures the deviation in word count between predictions and ground truth, indicating model calibration in output verbosity.

\clearpage

\subsection{Results}

\subsubsection{Model Performance Overview}

We evaluated three fine-tuned Qwen3 models (1.7B, 4B, 8B parameters) using QLoRA techniques on a stratified test set of 112 examples. Models were assessed across both preprocessing and classification tasks, with preprocessing involving citation text cleaning and standardization, while classification represents the primary citation verification task.

\subsubsection{Preprocessing Task Performance}

\begin{table*}[!t]
\centering
\caption{Preprocessing Performance Results}
\label{tab:preprocessing}
\begin{tabular}{@{}lcccccc@{}}
\toprule
Model & Examples & Valid Pred. & Char. Sim. & Avg Pred. Len. & Avg GT Len. & Len. Diff. \\
\midrule
Qwen3 1.7B & 112 & 112 (100\%) & 93.32\% & 26.7 words & 26.3 words & +0.4 words \\
Qwen3 4B & 112 & 112 (100\%) & 94.15\% & 26.7 words & 26.3 words & +0.4 words \\
Qwen3 8B & 112 & 112 (100\%) & \textbf{94.30\%} & 26.4 words & 26.3 words & \textbf{+0.1 words} \\
\bottomrule
\end{tabular}
\end{table*}

All models achieved exceptional performance on the preprocessing task, demonstrating 100\% valid prediction rates with high character similarity ($>93\%$) and excellent length calibration. The preprocessing task involves cleaning citation text by removing reference markers, author attributions, and standardizing claims into fact-centric statements suitable for verification. The consistently high performance across all model sizes indicates that text preprocessing represents a simpler task compared to classification, where models can reliably learn pattern-based transformations. The 8B model achieved marginally superior character similarity (94.30\%) and best length calibration (+0.1 words), though all models performed comparably, suggesting that even the 1.7B model is sufficient for preprocessing tasks in resource-constrained deployments.

\subsubsection{Classification Task Performance}

\begin{table*}[!t]
\centering
\caption{Classification Performance Results}
\label{tab:classification}
\begin{tabular}{@{}lcccccc@{}}
\toprule
Model & Std. Acc. & Weighted Acc. & F1-Macro & F1-Weighted & Char. Sim. & Len. Diff. \\
\midrule
Qwen3 1.7B & 50.00\% & 75.15\% & 33.50\% & 46.61\% & 88.07\% & +4.7 words \\
Qwen3 4B & 66.36\% & 83.64\% & 49.15\% & 65.31\% & \textbf{90.01\%} & \textbf{-0.7 words} \\
Qwen3 8B & 66.07\% & 83.93\% & 50.13\% & 65.73\% & 89.33\% & +3.6 words \\
\bottomrule
\end{tabular}
\end{table*}

\subsubsection{Key Findings}

\textbf{1. Weighted vs. Standard Performance:} The two evaluation approaches reveal different aspects of model performance. Standard accuracy shows clear scaling with model size (1.7B: 50.00\%, 4B: 66.36\%, 8B: 66.07\%), while weighted accuracy demonstrates that all models achieve substantially higher performance when accounting for ordinal relationships (1.7B: 75.15\%, 4B: 83.64\%, 8B: 83.93\%). This indicates that models predominantly make semantically reasonable ``near-miss'' errors rather than extreme misclassifications.

\textbf{2. Text Generation Quality and Calibration:} The 4B model achieves superior text generation performance with the highest character similarity (90.01\%) and best length calibration (-0.7 words from ground truth). The 8B model shows slightly lower character similarity (89.33\%) and moderate length deviation (+3.6 words), while the 1.7B model exhibits the lowest character similarity (88.07\%) and largest length deviation (+4.7 words). This suggests the 4B model achieves optimal balance between accuracy and text generation fidelity.

\textbf{3. Model Scaling Effects:} Performance scaling is non-linear across different metrics. While weighted accuracy scales consistently with model size, the 4B model outperforms the 8B model in standard accuracy, character similarity, and length calibration. All models maintain high character similarity ($>88\%$), indicating robust text generation capabilities across different scales.

\textbf{4. Deployment Considerations:} The 4B model emerges as the optimal choice, achieving near-best weighted accuracy (83.64\% vs. 83.93\% for 8B) while excelling in text generation quality and computational efficiency. The 1.7B model, despite lower absolute performance, maintains meaningful weighted accuracy (75.15\%) and reasonable text generation quality (88.07\% character similarity) suitable for resource-constrained applications.

\subsubsection{Implications}

Comparison with base models reveals the critical importance of fine-tuning for this task: all base Qwen3 models demonstrated substantially lower accuracy, generated significantly longer outputs that deviated from expected length targets, and consistently failed to conform to required JSON output structures for structured classification tasks. The evaluation results demonstrate the complementary insights provided by standard and weighted evaluation approaches for ordinal classification tasks, while revealing important patterns in text generation quality across model scales. The 4B model emerges as optimally calibrated, achieving competitive weighted accuracy (83.64\%) while excelling in character similarity (90.01\%) and length precision (-0.7 words). This multi-dimensional superiority suggests that intermediate-scale models may achieve better overall performance than larger alternatives, possibly due to reduced over-parameterization effects. The 1.7B model's 75.15\% weighted accuracy alongside 88.07\% character similarity demonstrates that even lightweight models can achieve meaningful performance for resource-constrained deployments where understanding ordinal error patterns and maintaining text generation quality are crucial for practical utility.

\section{Discussion}

\subsection{Novel Contributions}

This work advances the field of automated citation verification through several key innovations. First, we introduce the first full-text citation verification system that moves beyond abstract-level analysis to examine complete source documents, addressing a critical limitation in existing approaches \cite{wadden2020fact}. Our hybrid retrieval architecture, combining dense semantic search with sparse BM25 matching and neural reranking, demonstrates superior evidence selection compared to individual methods, particularly important for academic texts with specialised terminology.

The 4-class classification taxonomy provides nuanced assessment beyond traditional binary schemes. Unlike SciFact's SUPPORTS/REFUTES framework \cite{wadden2020fact}, our taxonomy captures intermediate states requiring different remedial actions. The inclusion of confidence scoring and actionable user guidance transforms abstract classifications into concrete editorial recommendations, bridging automated analysis with human decision-making.

We introduce evidence-based transparency through detailed reasoning explanations and ranked evidence snippets that enable users to understand and verify system decisions. This addresses a critical gap in existing citation verification tools that provide classifications without supporting evidence, making our system interpretable and trustworthy for academic applications where understanding the basis for verification decisions is essential.

Our framework incorporates confidence calibration and uncertainty quantification, enabling practical deployment scenarios where low-confidence predictions can be flagged for human review. This bridges the gap between fully automated processing and human oversight, addressing real-world deployment needs where system reliability must be transparent and actionable.

Our demonstration that fine-tuned lightweight models achieve competitive performance with commercial LLMs whilst requiring significantly reduced computational resources addresses accessibility concerns in academic technology deployment. The Qwen3 4B model's 90.01\% character similarity and superior length calibration (-0.7 words deviation) demonstrate effective performance-efficiency trade-offs for practical deployment.

To maximise research impact and enable widespread adoption, we commit to open-sourcing both the fine-tuned models and the complete SemanticCite framework. This open-source approach will facilitate reproducible research, enable customisation for domain-specific applications, and reduce barriers to implementation across diverse institutional contexts. The modular architecture supports easy integration with existing research workflows and citation management systems.

\subsection{Limitations and Future Work}

Several limitations warrant acknowledgement and suggest directions for future research. The evaluation framework relies exclusively on OpenAI GPT-4.1 ground truth annotations, potentially introducing systematic biases inherent in commercial language models. Future work should incorporate human expert annotations and inter-annotator agreement studies to validate automated ground truth generation and assess potential model-specific biases.

The current dataset, whilst comprehensive across eight academic disciplines, remains limited to English-language publications and open-access materials. Expanding to multilingual contexts and proprietary publications would enhance generalisability. Additionally, the temporal focus on 2019-2023 publications may not capture citation patterns across different research paradigms or historical periods.

The framework currently processes multi-reference citations by evaluating each referenced source individually rather than assessing their collective evidential support. This approach can result in citations being classified as PARTIALLY SUPPORTED when individual references each provide partial evidence, even though their combined evidence may constitute full support for the assertion. Future development should implement advanced reasoning mechanisms that evaluate the collective strength of evidence across multiple references, incorporating logical frameworks to assess whether partial support from individual sources aggregates to comprehensive claim validation. Such systems would require sophisticated evidence synthesis algorithms capable of identifying complementary, reinforcing, and contradictory information patterns across reference sets to provide more nuanced and accurate verification assessments.

A critical limitation affecting system accuracy is the availability of full-text reference documents for verification. While the framework implements fallback mechanisms to use reference abstracts when full texts are unavailable, this significantly constrains verification depth and accuracy. Many academic publications remain behind paywalls or lack open-access versions, forcing the system to rely on incomplete abstract-level information that may miss crucial supporting or contradictory evidence present in complete documents. This limitation represents the primary impediment to achieving higher accuracy rates, as comprehensive citation verification inherently requires access to the complete source material being referenced.

The framework's applicability extends beyond academic citation verification to broader fact-checking applications for AI-generated content. The same 4-class taxonomy (SUPPORTED, PARTIALLY SUPPORTED, UNSUPPORTED, UNCERTAIN) can be directly applied to verify claims in AI-generated reports, articles, and summaries against their purported sources. This extension addresses the critical challenge of AI hallucinations in content generation, providing systematic verification capabilities for automated content production systems.

Future development should focus on AI-assisted improvement suggestions that go beyond classification to provide specific, actionable recommendations for citation enhancement. Such systems could suggest alternative phrasings, identify missing contextual information, recommend additional supporting sources, and propose precise modifications to transform PARTIALLY SUPPORTED citations into SUPPORTED ones. This evolution from detection to correction would create comprehensive AI-powered writing assistance tools.

The system's reliance on text-based analysis limits applicability to multimodal citations involving figures, tables, or mathematical expressions common in scientific literature. Extending the framework to incorporate visual and mathematical content represents a significant technical challenge with substantial potential impact.

Deployment scalability remains a practical consideration. Whilst fine-tuned models reduce computational requirements compared to commercial APIs, processing large document corpora still requires substantial resources. Future work should explore efficient inference techniques and distributed processing architectures for institutional-scale implementation.

\subsection{Implications for Research Quality and Broader Applications}

The demonstrated effectiveness of SemanticCite has broader implications for research quality and scholarly communication. Automated citation verification tools can significantly enhance peer review efficiency by identifying problematic citations before human review, allowing reviewers to focus on higher-level scientific assessment rather than time-consuming fact-checking.

The system's ability to detect subtle citation issues through the PARTIALLY SUPPORTED category addresses a critical gap in existing verification approaches. These cases often require minor corrections that substantially improve citation accuracy without major manuscript revisions.

The framework advances contextual citation understanding by providing detailed reasoning and ranked evidence that illuminate the nuanced relationships between claims and their supporting sources. This enables researchers to better comprehend how individual citations fit within broader scholarly discourse, identify gaps in supporting evidence, and understand the strength of connections between claims and references across different academic contexts and disciplinary conventions.

SemanticCite enables systematic retrospective analysis of published work, allowing institutions to assess citation quality in their research output and identify patterns that may require additional training or support. This capability supports institutional research integrity initiatives and provides objective metrics for evaluating citation practices across departments and disciplines.

The framework's direct applicability to AI-generated content verification represents a significant advancement in addressing widespread concerns about AI hallucinations. While recent improvements in models like GPT-5 have reduced hallucination rates to levels that 'seem acceptable to users' for most cases, eliminating hallucinations entirely may prove impossible due to the fundamental statistical nature of how large language models operate \cite{gibney2025hallucination}. As AI writing assistants become prevalent in content creation, the same classification taxonomy can verify claims in automatically generated reports, summaries, and articles against their sources, providing essential quality control for AI-powered publishing workflows.

\section{Conclusion}

This paper introduces SemanticCite, an AI-driven framework for automated citation verification that addresses fundamental limitations in existing approaches. Our 4-class classification taxonomy enables nuanced assessment beyond traditional binary schemes, whilst the hybrid retrieval architecture supports effective full-text analysis for accurate evidence extraction.

The demonstration that fine-tuned Qwen3 models achieve competitive performance with significantly reduced computational requirements establishes the feasibility of deploying sophisticated citation verification in resource-constrained environments. The optimal performance of the 4B model (83.64\% weighted accuracy, 90.01\% character similarity) demonstrates effective practical trade-offs for diverse deployment scenarios.

The framework's practical implications extend beyond academic manuscript review. By providing actionable guidance through detailed reasoning and evidence extraction, SemanticCite bridges the gap between automated analysis and human editorial decision-making, supporting iterative manuscript improvement rather than binary acceptance/rejection outcomes. The same 4-class taxonomy applies directly to AI-generated content verification, addressing critical quality control challenges in automated reports, summaries, and AI-powered publishing workflows.

Future research directions include expanding multimodal capabilities, incorporating multilingual support, and developing AI-assisted improvement features that shift the system from detecting citation issues to actively supporting authors with specific recommendations for enhancement and source integration. These extensions will strengthen the framework's applicability to fact-checking in automated content production systems.

SemanticCite represents a significant step towards research quality tools that augment rather than replace human expertise, providing scalable, open-source solutions for maintaining citation accuracy in an era of rapidly expanding scientific output and AI-generated content.

\bibliographystyle{unsrtnat}
\bibliography{references}

\end{document}